\documentclass[a4paper]{article}

\usepackage{INTERSPEECH2020}
\usepackage{graphicx,xcolor}

\title{Serialized Output Training for End-to-End Overlapped Speech Recognition}
\name{Naoyuki Kanda, Yashesh Gaur, Xiaofei Wang, Zhong Meng, Takuya Yoshioka}
\address{
  Microsoft Corp.} 
\email{\{Naoyuki.Kanda,Yashesh.Gaur,Xiaofei.Wang,Zhong.Meng,tayoshio\}@microsoft.com}

\begin{document}

\maketitle
\begin{abstract}
This paper proposes serialized output training (SOT), 
a novel framework for multi-speaker overlapped speech recognition
based on an attention-based encoder-decoder approach. 
Instead  of  having  multiple  output  layers  as with the permutation invariant training (PIT), SOT uses a model with only one output layer that generates the  transcriptions  of  multiple speakers  one  after  another. 
The attention and decoder modules take care of producing 
multiple transcriptions from overlapped speech. 
SOT has two advantages over PIT: 
(1) no limitation in the maximum number of speakers, and
(2) an ability to model the dependencies among outputs for different speakers.
We also propose a simple trick 
that allows SOT to be executed 
in $O(S)$, where $S$ is the number of the speakers in the training sample, by using the start times of the constituent source utterances. 
Experimental results on LibriSpeech corpus show that the SOT models can transcribe overlapped speech with variable numbers of speakers 
significantly better than PIT-based models.
We also show that the SOT models can accurately count the number of speakers in the input audio. 
\end{abstract}
\noindent\textbf{Index Terms}: multi-speaker speech recognition, attention-based encoder-decoder, permutation invariant training, serialized output training

\section{Introduction}

Thanks to the advancement in deep neural network (DNN)-based automatic speech recognition (ASR) \cite{seide2011conversational,hinton2012deep},
the word error rate (WER) for single speaker 
recordings has 
reached to
the level of human transcribers even for tasks
that were once thought very challenging (e.g., Switchboard \cite{xiong2016achieving,saon2017english}, LibriSpeech \cite{amodei2016deep,povey2016purely,kanda2018lattice,luscher2019rwth,park2019specaugment,han2019state}) .
Nonetheless, ASR for multi-speaker speech remains to be a very difficult problem
especially when multiple utterances significantly overlap in monaural recordings. 
For example, an ASR system that achieves a WER of 5.5\% for single speaker speech can yield a WER of 84.7\% 
for two-speaker overlapped speech as reported in \cite{kanda2019auxiliary}.

Researchers have made tremendous efforts towards 
the multi-speaker ASR for handling overlapped speech.
One of the early works with DNN-based ASR is 
to train two ASR models, one of which recognizes a speech with higher instantaneous energy and
another one of which recognizes a speech with lower instantaneous energy \cite{weng2015deep}.
This method has a limitation that the model can handle only two speakers.
A more sophisticated method for multi-speaker ASR is the permutation invariant training (PIT)
in which the model has multiple output layers corresponding to different speakers,
and the model is trained by considering all possible permutations of
speakers.
PIT was 
proposed for speech separation \cite{yu2017permutation} 
and 
multi-speaker ASR \cite{yu2017recognizing},
and worked well for both of them.
Despite this, 
PIT has several limitations.
Firstly, the number of the output layers in the model
constrains the maximum number of speakers that the model can output.
Secondly, it cannot handle the dependency among utterances of multiple speakers because 
the output layers are independent from each other.
Because of this, it is possible that the duplicated hypotheses are generated 
from different output layers, and extra treatment is necessary
to reduce such errors \cite{seki2018purely}.
Thirdly, 
the computational complexity of PIT is at the order of $O(S^3)$,
where $S$ represents
the number of speakers. 
Because of these limitations, most previous works using PIT \cite{yu2017permutation,yu2017recognizing,seki2018purely,chang2019end,chang2019mimo} only addressed the two-speaker case
although real recordings often contain even more speakers.

In this paper, we propose a novel framework for overlapped speech recognition,
named Serialized Output Training (SOT),
based on 
an attention-based encoder-decoder (AED) approach \cite{bahdanau2014neural,chorowski2014end,chorowski2015attention,chan2016listen}. 
Instead of having multiple output layers as with the PIT-based ASR, 
our proposed model has only one output layer
that generates the transcriptions of multiple speakers
one after another.
By using the single output layer for the modeling, we avoid having the maximum speaker number constraint.
In addition, the proposed method can naturally model the dependency among the outputs for multiple speakers,
which could help 
avoid duplicate hypotheses from being generated. 
We also propose a simple trick 
that allows SOT to be executed
in $O(S)$ 
by using the start times of the 
source utterances.
We show that the proposed method can 
better 
transcribe 
utterances of multiple speakers 
from monaural overlapped speech
than PIT, and can 
count 
the number of speakers with good accuracy.

\begin{figure*}[t]
  \centering
  \includegraphics[width=\linewidth]{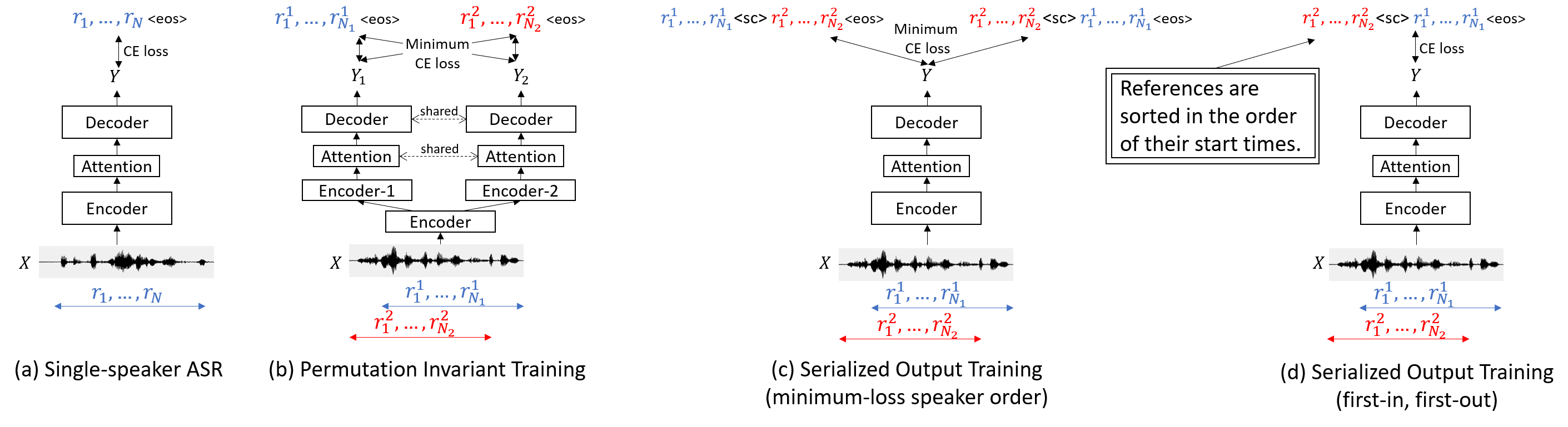}
  \vspace{-8mm}
  \caption{Architectures of (a) the conventional single-speaker ASR, (b) the conventional multi-speaker ASR using PIT, and (c),(d) the proposed serialized output training (SOT) with different schemes. The special symbol $\langle sc\rangle$ represents the speaker change, and is inserted in between the utterances. }
  \label{fig:asr_models}
  \vspace{-5mm}
\end{figure*}

\section{Related Work}
The most relevant work we are aware of would be the joint speech recognition and diarization with 
a recurrent neural network transduer (RNN-T) \cite{el2019joint}. In the paper, the authors 
proposed to generate transcriptions of different speakers interleaved by speaker role tags to recognize two-speaker conversations. 
Another related piece of work is the AED-based multilingual ASR for mixed-language speech~\cite{watanabe2017language,seki2018end}.
They used language tags as additional output symbols to transcribe the mixed-language speech with a single model.
These methods are similar to ours in that both approaches decode multiple utterances that are separated by a special symbol. 
However, all the aforementioned methods did not deal with speech overlaps.

It should be noted that,
although AED was originally proposed for machine translation to cope with word order differences  
between the source and target languages \cite{bahdanau2014neural},
the previous studies on the AED-based ASR 
attempted to incorporate a monotonic alignment constraint
to reduce errors in attention estimation. 
For example, 
\cite{chorowski2014end} used a penalty for encouraging monotonic alignment, and 
\cite{kim2017joint} proposed jointly training 
connectionist temporal classification (CTC) and AED models. 
Other popular ASR frameworks, such as the hybrid of DNN and hidden Markov model (HMM), CTC, and RNN-T,
also impose the monotonic alignment assumption.
By contrast, 
our attention module scans the encoder embedding sequence back and forth along the time dimension
to transcribe utterances of multiple speakers, 
which is the key difference from the previous studies.

\section{Review: Multi-Speaker ASR Based on AED with PIT}
\subsection{AED-based single-speaker ASR}

The AED-based ASR consists of encoder, attention, and decoder modules as shown in Fig. \ref{fig:asr_models} (a). Given input $X=\{x_1,...,x_T\}$, the AED produces 
the output sequence $Y=\{y_1, ..., y_n, ...\}$ as follows. 
Firstly, the encoder converts the input sequence $X$ 
into a sequence, $H^{enc}$, of embeddings, i.e.,  
 \begin{align}
 H^{enc} &=\{h^{enc}_1,...,h^{enc}_T\}={\rm Encoder}(X).  \label{eq:enc} 
\end{align}
Then, for every decoder step $n$, the attention module 
outputs context vector $c_n$ with attention weight $\alpha_n$
given decoder state vector $q_n$, the previous attention weight $\alpha_{n-1}$, and
the encoder embeddings $H^{enc}$ as 
\begin{align}
 c_n, \alpha_n &= {\rm Attention}(q_n, \alpha_{n-1}, H^{enc}). \label{eq:att} 
\end{align}
Finally, 
the output distribution $y_n$ 
is estimated given
the context vector $c_n$
and the decoder state vector $q_n$ as follows: 
\begin{align}
 q_n &={\rm DecoderRNN}(y_{n-1}, c_{n-1}, q_{n-1}), \\
 y_n&= {\rm DecoderOut}(c_n,q_n). \label{eq:dec}
\end{align}
Here, 
${\rm DecoderRNN}$ consists of multiple RNN layers while ${\rm DecoderOut}$ consists of
an affine transform with a softmax output layer.
The model is trained to minimize the cross entropy loss
given $Y$ and reference label $R=\{r_1,...,r_N, r_{N+1}=\langle eos\rangle\}$. Specifically, 
the loss function is defined as 
\begin{align}
    \mathcal{L}^{CE}=\sum_{n=1}^{N+1}{\rm CE}(y_n, r_n), 
\end{align}
where
${\rm CE()}$ represents the cross entropy given the output distribution
and the reference label,
and $N$ is the number of symbols in the reference $R$.
$\langle eos\rangle$ is the special symbol that represents the end of the sentence.

\subsection{PIT-based ASR with multiple output layers}

With the conventional multi-speaker ASR with PIT, 
the model has multiple output branches 
as shown in Fig. \ref{fig:asr_models} (b).
Thus, we have 
\begin{align}
H^{enc_s}&={\rm Encoder}_s(H^{enc})\\
 c^s_{n}, \alpha^s_{n} &= {\rm Attention}(q^{s}_{n}, \alpha^s_{n-1}, H^{enc_s}) \\
 q^s_{n} &= {\rm DecoderRNN}(y^s_{n-1}, c^s_{n-1}, q^s_{n-1}),\\
 y^s_n&= {\rm DecoderOut}(c^s_n,q^s_n). \label{eq:decpit}
\end{align}
Here, $s$ is the index of each output branch, where $1 \leq s \leq S$ with $S$ being the number of speakers. 
Parameters for the attention
and decoder modules are shared across $s$.
Given the set of the outputs, $\{Y^1, \cdots, Y^S\}$, and 
the set of the references, $\{R^1, \cdots, R^S\}$, where $R^s$ denotes the $s$th reference defined as  $R^s=\{r^s_{1},..,r^s_{N^s}, r^s_{N^s+1}=\langle eos\rangle\}$,
the PIT-CE loss is calculated 
by considering all 
possible speaker permutations as
\begin{align}
    \mathcal{L}^{PIT}=\min_{\phi\in 
    {\Phi}(1,...,S)}\sum_{s=1}^{S}\sum_{n=1}^{N^s+1}{\rm CE}(y^{\phi[s]}_n, r^s_{n}). 
\end{align}
Here, ${\Phi()}$ is the function that generates all possible permutations of a given sequence.

There are three theoretical limitations in PIT. 
Firstly, the number of the output layers in the model
constrains the maximum number of speakers that the model can handle.
Secondly, it cannot represent the dependency between the utterances of multiple speakers because 
each output $Y^s$ has no direct dependency on the other outputs. 
Because of this, it might be possible that the duplicated hypotheses are generated from each output layer.
Thirdly, 
even with the Hungarian algorithm \cite{kuhn1955hungarian},
it requires a training cost of $O(S^3)$ 
which hinders its application to a large number of speakers.

\section{Serialized Output Training}

\subsection{Overview}
Instead of having multiple output layers as with PIT,
we propose to use the original form of AED (Eq. \eqref{eq:enc}-\eqref{eq:dec}), which has only one output branch, 
for the multi-speaker ASR. 
To recognize multiple utterances, 
we serialize multiple references into a single token sequence. 
Specifically, 
we introduce 
 a special symbol $\langle sc\rangle$ to represent the speaker change 
and simply concatenate the reference labels 
by inserting 
 $\langle sc\rangle$
 between utterances.
For example, for a three-speaker case, the reference label will be given as
$R=\{r^1_{1},..,r^1_{N^1}, \langle sc\rangle, r^2_{1},..,r^2_{N^2}, \langle sc\rangle, r^3_{1},..,r^3_{N^3}, \langle eos\rangle\}$.
Note that $\langle eos\rangle$ 
is used 
only at the end of the entire sequence. 
We call our proposed approach serialized output training (SOT).

Because there are multiple permutations in the order of reference labels
to form $R$, 
some trick is needed to calculate the loss 
between the
output $Y$ and the concatenated reference label $R$.
For example, in the case of two speakers, the reference label can be either
$R=\{r^1_{1},..,r^1_{N^1}, \langle sc\rangle, r^2_{1},..,r^2_{N^2}, \langle eos\rangle\}$
or $R=\{r^2_{1},..,r^2_{N^2}, \langle sc\rangle, r^1_{1},..,r^1_{N^1}, \langle eos\rangle\}$.
One possible way 
to determine the order 
is 
to calculate the loss for all possible concatenation patterns to form $R$
and select the best one, similarly to PIT, as (Fig. \ref{fig:asr_models} (c))
\begin{align}
        \mathcal{L}^{SOT-1}=\min_{\phi\in 
        {\Phi}(1,...,S)}\sum_{n=1}^{N^{sot}}{\rm CE}(y_n, r^\phi_{n}),
\end{align}
where $N^{sot}=\sum_{s=1}^S \{N^s+1\}$, and $r^\phi_{n}$ is the $n$-th token in 
the concatenated reference given permutation $\phi$.
We call this method as SOT with minimum-loss speaker order. 
This method has the problem that requires $O(S!)$ training cost. 

To reduce the computational cost to $O(S)$, 
we alternatively propose to sort the reference labels
by their start times (Fig. \ref{fig:asr_models} (d)) as follows:
\begin{align}
\mathcal{L}^{SOT-2}=
\sum_{n=1}^{N^{sot}}{\rm CE}(y_n, r^{\Psi(1,...,S)}_{n}).
\end{align}
Here, $\Psi$ is the function that outputs the sorted index of $\{1,...,S\}$ 
according to the start time of each 
speaker. The term $r^{\Psi(1,...,S)}_{n}$ is the $n$-th token in the concatenated reference
given $\Psi(1,...,S)$.
As a result, the AED is trained to recognize the utterances of 
multiple speakers in the order of their start times, separated by 
a special symbol $\langle sc\rangle$.
We call this method as SOT based on first-in, first-out order\footnote{A similar idea to use the start times was proposed in \cite{tripathi2020end} during this paper was reviewed.}.
The only assumption to perform this first-in, first-out training is that the
two utterances do not start at exactly the same time.
If that is the case, we randomly determine the utterance order. 
That said, it rarely happens in real recordings, and thus its impact should be marginal.

\subsection{Separation after attention (SAA)}
\label{sec:sep-after-attention}

With PIT, speech separation is explicitly modeled by the multiple encoder branches.
In the proposed SOT framework,
the attention module 
operates on the encoder embeddings that could be contaminated by overlapped speech.
Thus, 
the context vector generated by the attention module may also be contaminated, resulting in potential degracation in accuracy.

We found that simply inserting 
one unidirectional LSTM layer in DecoderOut() in Eq \eqref{eq:dec} 
could solve the problem effectively. 
Unlike PIT, where the speech separation is performed {\it before} the attention module, 
this additional LSTM works as a
separation module, taking place {\it after} the attention module.
In our experiment, we removed one encoder layer  when we applied this ``Separation after Attention (SAA)'' method for the sake of fairness in terms of the model size.

\subsection{Advantages of the proposed method}

There are two key advantages of using single output branch instead of
multiple branches as with PIT.
Firstly, 
 there is no longer a limitation on the maximum number of speakers
 that the model can handle.
Secondly, the proposed model can 
represent the dependency among multiple utterances, which 
prevents duplicated hypotheses from being generated.

Furthermore, 
the proposed model can even predict the number of speakers 
if the model is trained on a data set including various numbers of speakers.
At the inference time, the decoder module 
will not stop until $\langle eos\rangle$
is predicted. 
Therefore, it can automatically count the number of speakers 
in the recording just by counting the occurrences of 
$\langle sc\rangle$ and $\langle eos\rangle$.

\section{Experiments}
\subsection{Evaluation settings}

\subsubsection{Training and evaluation data}
In this work, we used the LibriSpeech corpus \cite{panayotov2015librispeech} to simulate multi-speaker signals and evaluate the proposed method.
LibriSpeech consists of about 1,000 hours of audio book data.
We followed the Kaldi \cite{povey2011kaldi} recipe to generate the dataset,
and used
the 960 hours of training data ("train\_960") for training the ASR models,
and 
used "dev\_clean" and "test\_clean" for the evaluation.

Our training data were generated as follows. 
For each training example, 
we firstly determined the number of speakers $S$ to be included in the sample.
For each utterance in "train\_960", 
$(S-1)$ utterances were randomly picked up from "train\_960"
and added with random delays. 
When mixing the audio signals, the original volume of each utterance was kept unchanged, resulting in an average signal-to-interference ratio of about 0 dB. 
As for the delay applied to each utterance, the delay values were randomly chosen under the constraints 
that (1) the start times of the individual utterances differ by 0.5 s or longer
and that (2) every utterance in each mixed audio sample has at least one speaker-overlapped region with other utterances.

The evaluation set was generated from "dev\_clean" or "test\_clean" in the same way except that Constraint (1) mentioned above was not imposed.  
Therefore,
multiple utterances were allowed to start at the same time in the evaluation data.

\subsubsection{Evaluation metric}
Our trained models were evaluated with respect to WER. 
In multi-talker ASR, a system may produce a different number of hypotheses than references (i.e., speakers). 
To cope with this, all possible permutations of the hypothesis order were examined, and the one that yielded the lowest WER was picked.

\begin{table}[t]
  \caption{WER(\%) of 512-dim models for {\bf 2-speaker-mixed speech}. 
  Note that the WERs for single speaker speech by the single-speaker ASR were 5.4\% and 5.7\% for dev\_clean and test\_clean, respectively. } 
  \label{tab:base}
  \vspace{-3mm}
  \centering
  {\footnotesize
  \begin{tabular}{c|cc}
    \toprule
Model & \multicolumn{2}{c}{WER (\%)} \\
{\scriptsize (all 512-dim)}                 & dev\_clean & test\_clean  \\ \midrule
Single-speaker ASR  &  67.9   &  68.5 \\ \midrule
SOT (minimum loss speaker order) &  17.4     &  17.1 \\
    SOT (first-in, first-out) &17.0  & 16.5 \\
    \bottomrule
  \end{tabular}
  }
  \vspace{-5mm}
\end{table}

\subsubsection{Model settings}
In our experiments,
we used 6 layers of $M$-dim bidirectional long short-term memory (BLSTM) for the encoder,
where $M$ was set to 512, 724 or 1024. 
Layer normalization \cite{ba2016layer} was applied after every BLSTM.
For PIT-based baseline systems, the first 5 encoder layers were shared
across the output branches, and each branch had its own last encoder layer.
The decoder module consists of 
2 layers of $M$-dim unidirectional LSTM without layer normalization.
We used a conventional location-aware content-based attention \cite{chorowski2015attention} with 
a single head. 

As for the input feature, 
we used an 80-dim log mel filterbank extracted every 10 msec.
We stacked 3 frames of features and applied the encoder on top of the stacked features.
We used 16k subwords based on a unigram language model \cite{kudo2018subword}
as a recognition unit. 
We applied the speed and volume perturbation \cite{ko2015audio} to the mixed audio to enhance the model training.

We used the Adam optimizer with a learning rate schedule similar to that described in \cite{park2019specaugment}.
We firstly linearly increased the learning rate from 0 to 0.0002 by using the initial 1k iterations and kept the learning rate until the 160k-th iteration.
We then started decaying the learning rate exponentially at a rate of  1/10 per 240k iterations.
In this paper, we report the results of the``dev\_clean''-based best models found after 320k of training iterations. 
We used minibatch consists of 9k, 7.5k, and 6k frames of input for $M$=512, 724, 1024, respectively.
8 GPUs were used for all training.

Note that
we applied neither an additional language model nor 
SpecAugment 
\cite{park2019specaugment}
for simplicity. 
Our results
for the standard LibriSpeech ``test\_clean'' 
(4.6\% of WER in Table \ref{tab:pit-vs-sot}) 
was on par with
recently reported results without these techniques
(eg, 4.1\%--4.6\% of WER was reported in
\cite{luscher2019rwth,park2019specaugment,irie2019language,zeyer2019comparison,rosenberg2019speech}).

\begin{table}[t]
  \caption{WER (\%) of 512-dim models for {\bf various mixtures} of training and test data. Test data were generated by mixing ``test\_clean''. Evaluation results with unmatched training/testing conditions were shown with parenthesis.} 
  \label{tab:variable-speaker}
  \vspace{-3mm}
  \centering
  {\footnotesize
  \begin{tabular}{cc|ccc}
    \toprule
Model & \# of Speakers & \multicolumn{3}{c}{\# of Speakers in Test Data} \\
{\scriptsize (all 512-dim)} & in Training Data     & 1 & 2 & 3 \\
    \midrule
Single-spk ASR   & 1 & 5.7 & (68.5) & (92.7)\\ \midrule
SOT (fifo) &2 & (16.7) & 16.5  & (55.2) \\
SOT (fifo)  & 1,2  & 5.7  & 18.6 & (59.7) \\
SOT (fifo)  & 1,2,3  &  5.4 & 17.3 &  34.3 \\ 
    \bottomrule
  \end{tabular}
  }
  \vspace{-3mm}
\end{table}

\begin{table}[t]
  \caption{WER (\%) for different numbers of parameters and architectures for SOT model trained with the mixture of 1, 2, and 3 speakers. Test data were generated by mixing ``test\_clean.''} 
  \label{tab:modelsize}
  \vspace{-3mm}
  \centering
  {\footnotesize
  \begin{tabular}{ccc|ccc|c}
    \toprule
Model&  SAA & \# of & \multicolumn{4}{c}{\# of Speakers in Test Data} \\
Dim &  (Sec \ref{sec:sep-after-attention})& Params & 1 & 2 & 3 & {\bf Total}\\
    \midrule
512 & & 44.7M    & 5.4 & 17.3 & 34.3 & {\bf 23.8} \\ 
724 & & 79.2M & 4.5 & 12.0   & 26.5 & {\bf 18.0}\\ 
1024  & &        143.9M & 4.8 & 10.9 & 25.8 & {\bf 17.3}\\ \midrule
1024 & $\surd$ & 135.6M & 4.6  & 11.2 & 24.0 & {\bf 16.5} \\
    \bottomrule
  \end{tabular}
  }
  \vspace{-5mm}
\end{table}

\subsection{Evaluation results}
\subsubsection{Evaluation on two-speaker mixed speech}

First, we
evaluated the SOT model for two-speaker overlapped speech.
In this experiment, 
both the training and evaluation data consisted only of
two-speaker overlapped utterances.
We used a 512-dim AED model without SAA.

As shown in Table \ref{tab:base}, the SOT model significantly outperformed a normal single-speaker ASR.
Surprisingly,
the ``first-in, first-out'' training achieved
a better WER than the ``minimum loss speaker order'' training
although the former approach can be
executed with a computational cost of $O(S)$ 
with respect to the number of speakers.
Based on this finding, we used the ``first-in, first-out'' training for the rest of the experiments.

\subsubsection{Evaluation on variable numbers of overlapped speakers}
We then evaluated the same 512-dim model with a training set consisting of various numbers of speakers.
The results are shown in Table \ref{tab:variable-speaker}, where
the SOT model is shown to be able to recognize overlapped speech with variable numbers of speakers very well.
It should be emphasized that the SOT model 
did not show any degradation for the single-speaker case, 
sometimes
even outperformed the single-speaker ASR model. 
This might be because of the
data augmentation effect resulting from mixing training utterances.

We also investigated the impact of the number of parameters,
the results of which are shown in Table \ref{tab:modelsize}.
We found that the large model size was essential for SOT to achieve good results. 
This is because the attention and decoder modules in the SOT framework are required to work
on contaminated embeddings as we discussed in Sec \ref{sec:sep-after-attention}.
Applying SAA further improved the performance
as shown in the last row of Table \ref{tab:modelsize}.
Note that the SOT model size of with SAA 
is smaller than
the naive SOT because we removed one 1024-dim {\it bidirectional} LSTM layer from the encoder instead of simply inserting one 1024-dim
{\it unidirectional} LSTM after the attention module.

\begin{table}[t]
  \caption{WER (\%) comparison of PIT and SOT. Test data were generated by mixing ``test\_clean''.
  Evaluation results with unmatched training/testing conditions were shown with parenthesis.
 Note that we trained the PIT model 
  with up to 2 speakers since 3-output PIT required 
  impractical training time.} 
  \label{tab:pit-vs-sot}
  \vspace{-3mm}
  \centering
{\scriptsize
  \begin{tabular}{ccc|ccc}
    \toprule
 Model &{\scriptsize\# of Speakers} & \# of & \multicolumn{3}{c}{{\scriptsize\# of Speakers in Test Data}} \\
{\scriptsize (1024-dim)} & {\scriptsize in Training Data} & Params.    & 1 & 2 & 3 \\
    \midrule
2-output PIT  & 2  & 160.7M & (80.6) & 11.1 & (52.1) \\ 
2-output PIT  & 1,2  & 160.7M & 6.7 & 11.9 & (52.3) \\  \midrule
SOT & 1,2,3  & 135.6M &  4.6 &  11.2 &  24.0 \\
    \bottomrule
  \end{tabular}
  }
  \vspace{-3mm}
\end{table}

\begin{table}[t]
  \caption{Speaker counting accuracy (\%) for the 1024-dim SOT model trained by the mixture of 1, 2, 3 speakers.}
  \label{tab:spk-count}
  \vspace{-3mm}
  \centering
  {\scriptsize
  \begin{tabular}{c|cccc}
    \toprule
Actual \# of Speakers& \multicolumn{4}{c}{Estimated \# of Speakers (\%)} \\
in Test Data & 1 & 2 & 3 & $>$4 \\
    \midrule
1 & {\bf 99.8} & 0.2 & 0.0 & 0.0\\ \midrule
2 & 1.9 & {\bf 97.0} & 1.1 & 0.0 \\ \midrule
3 & 0.2& 24.0 & {\bf 74.2} & 1.5\\ 
    \bottomrule
  \end{tabular}
  }
  \vspace{-5mm}
\end{table}

\subsubsection{Comparison with PIT}
\label{sec:pit-vs-sot}
Table \ref{tab:pit-vs-sot} compares SOT with PIT. 
In this experiment,
we used 1024-dim model for both PIT and SOT.
As shown in the table,
PIT showed severe WER degradation for the 1-speaker case
even when 
the training data contained many 1-speaker examples. 
SOT achieved significantly better results than the PIT model
trained on 1- and 2-speaker mixtures for both 1- and 2-speaker evaluation data
while recognizing 3-speaker evaluation data, with fewer parameters.
For reference, the training speed for the 3-speaker SOT was roughly 30\% faster (including I/O) than the 2-output PIT. The difference of the training speed would become much larger if we use PIT with more
output branches.

\subsubsection{Speaker counting accuracy}
Finally, we analyzed the speaker counting accuracy of the 1024-dim SOT.
As shown in Table \ref{tab:spk-count},
we found that the model could count the number of speakers very accurately especially for 
1-speaker (99.8\%) and 2-speaker cases (97.0\%) while it sometimes underestimated the speakers for the 3-speaker mixtures.

\section{Conclusions}
In this paper, 
we proposed SOT that can recognize overlapped speech consisting of any number of speakers.
We also proposed
a simple trick to 
execute SOT
in $O(S)$
by using the start time of each utterance. 
Our experiments on LibriSpeech showed
that the proposed model could transcribe utterances 
from monaural overlapped speech
significantly more effectively than PIT while
being able to accurately count the number of speakers as well.

\bibliographystyle{IEEEtran}

\bibliography{mybib}

\end{document}